%% file: main.tex
\input{kth_thesis_style_papers}

\usepackage[headheight=15pt]{geometry}  
\usepackage[intlimits]{mathtools}       
\usepackage{graphicx}                   
\usepackage{booktabs}                   
\usepackage{multirow}                   
\usepackage[normalem]{ulem}             
\useunder{\uline}{\ul}{}                
\usepackage{appendix}                   
\usepackage{xspace}
\usepackage{amsthm}                     
\usepackage{mathrsfs}                   
\usepackage{amssymb}                    
\usepackage{hyperref}                   
\usepackage{natbib}                     
\usepackage{lscape}                     
\usepackage{mathtools}                  
\usepackage{algpseudocode}              
\usepackage[Algorithmus]{algorithm}     
\hypersetup{
    pdftitle={Population Synthesis using Incomplete Information},
    pdfauthor={Tanay Rastogi},
    colorlinks=true,
    linkcolor=black,
    citecolor=black,
    filecolor=black,
    urlcolor=black
}

\begin{document}

\title{Population Synthesis using Incomplete Information}
\author{Tanay Rastogi, Daniel Jonsson and Anders Karlström}
\date{}
\maketitle
\pagenumbering{arabic}

\begin{abstract}
\noindent This paper presents a population synthesis model that utilizes the Wasserstein Generative-Adversarial Network (WGAN) for training on incomplete microsamples. By using a mask matrix to represent missing values, the study proposes a WGAN training algorithm that lets the model learn from a training dataset that has some missing information. The proposed method aims to address the challenge of missing information in microsamples on one or more attributes due to privacy concerns or data collection constraints. The paper contrasts WGAN models trained on incomplete microsamples with those trained on complete microsamples, creating a synthetic population. We conducted a series of evaluations of the proposed method using a Swedish national travel survey. We validate the efficacy of the proposed method by generating synthetic populations from all the models and comparing them to the actual population dataset. The results from the experiments showed that the proposed methodology successfully generates synthetic data that closely resembles a model trained with complete data as well as the actual population. The paper contributes to the field by providing a robust solution for population synthesis with incomplete data, opening avenues for future research, and highlighting the potential of deep generative models in advancing population synthesis capabilities.
\end{abstract}


\newpage
\section{Introduction}\label{Introduction}
Transportation simulation models, using agent-based models (ABMs), are widely used for various tasks like predicting travel demand, evaluating policy impacts or analyzing travel behavior. These models usually need complete information about individuals' social and household characteristics from an area covering cities, towns, or even countries (\cite{Bastarianto2023Agent-basedOpportunities}). Ideally, such information could be collected from census data at an individual or household level, and then we could draw a certain number of samples as a synthetic population. Statistical authorities in many countries have also made available a microsamples of individual-level data from the whole population and can be used in place of census data. Travel surveys that capture complete demographic and socioeconomic attributes at a comparable sampling rate can also act as a good replacement. In addition to these microsamples, aggregated marginal information on a regional or zonal level is usually available from the Bureau of Statistics. However, acquiring granular, individual-level data is challenging. Issues such as privacy concerns, as well as the technical and financial constraints of data gathering, often impede accessibility to detailed data. 

In order to tackle difficulties in data collection, population synthesis algorithms were created to create synthetic populations. They provide ABM transport models with a reliable alternative to actual populations. Population synthesis techniques generate a comprehensive list of a simulated population, each accompanied by corresponding attribute data. The objective of population synthesis is to optimally utilize the existing microsamples, along with the additional aggregated or marginal information on each attribute of interest, in order to generate agents that closely aligns with the underlying population structure (\cite{Sun2015ASynthesis}). These simulated agents can thereafter be employed to evaluate the impact of factors such as governmental policies on the region or conduct studies that would be prohibitively costly, ethically questionable, or just unfeasible using actual population data.

As highlighted in the research by \cite{Rich2018Large-scaleDenmark, Borysov2019HowSynthesis} the population synthesis methods, typically consist of three steps - 1). Starting solution where a synthetic pool of individuals are generated to represent a diverse combinations of attributes. This is usually done based on the available microsamples. 2). Fitting stage where weighting factors for the synthetic pools are estimated to construct the representative synthetic population for future targets. 3). Allocation stage where synthetic agents are generated and are assigned to ABM transport models. In this paper, we focus exclusively on the first stage—the generating of appropriate representative samples for a given population without considering how such samples can be aligned with future targets. This step is crucial because any underlying error in this step will propagate in the ABM models and hence affect the forecast results. 

In the population synthesis literature, there has been a trend toward simulation based probabilistic models for the first stage i.e. generation of starting solution. Research from \cite{Farooq2013SimulationSynthesis, Sun2015ASynthesis, Borysov2019HowSynthesis, Garrido2020PredictionModelling, Kim2023ASynthesis} utilized simulation-based generative models to create synthetic populations for the first stage of population synthesis. One crucial similarity in each of these studies is that the microsample used in the experiments are from single data-source and are complete i.e. there is no missing information for any attributes in the microsample. The quality of generated synthetic populations depends, of course, on the quality and detail of the microsample. While the data quality has been improving, it has not kept pace with the growing interest in microsimulations at the scale of individuals tagged with many associated attributes. Available microsamples are often thin and incomplete. Survey microsample cannot comprehensively cover all the variations of different attributes found in the actual population. Frequently, these microsamples exhibit incomplete data on one or more attributes due to errors in data collection or the respondents intentionally withholding information or not collecting it to ensure privacy. Additionally, to improve the attribute richness of the microsample, they can be combined with supplemental data from other sources. For example, travel survey in a certain region from multiple distinct organizations can be combined to add more data and attributes in the microsample. However, there may be situations in which one or more attributes are absent in either of the surveys. Hence, there is a need for an imputation algorithm can be used to estimate missing values based on data that was observed/measured in microsample. This will ensure that the synthetic population generated by population synthesis models is complete.

In this context we propose a new approach that uses the incomplete microsamples in order to draw the synthetic populations from it. In the context of this study, the \textit{incomplete microsamples} is one which has missing info on one or more attributes of the sample. We use a population synthesis model based on the Wasserstein Generative Adversarial network (WGAN) suggested by \cite{Kim2023ASynthesis}. The contribution of this study lies in the proposal of a novel technique in the WGAN training algorithm that enables the model to effectively learn using incomplete training data. The aim of this study is to propose a methodology that can effectively addresses the challenges associated with missing data, while ensuring a degree of accuracy that is at least equivalent to the methods described in the existing literature.

The remainder of the article is structured in the following manner: \textbf{\textit{Sec.}} \ref{References} provides a overview of the existing literature on the subject matter and highlights the contribution of this paper within the field. The proposed training methodology for the WGAN model is presented in \textbf{\textit{Sec.}} \ref{Method}. We first briefly present the original WGAN training method and then describe the proposed changes to it. To evaluate and access the performance of the proposed training method, we utilize a microsample from a Swedish national travel survey. The experimental setup, metric evaluation, results, and discussions on these are provided in \textbf{\textit{Sec.}} \ref{sec:CaseStudy}. Finally, in \textbf{\textit{Sec.}} \ref{sec:Conclusion}, the article concludes by summarizing the analysis and suggesting potential avenues for future research.

\section{Literature review}\label{References}

As proposed by \cite{Sun2018ASynthesis, Borysov2019HowSynthesis} the population synthesis methods can be divided into two primary categories: deterministic and simulation methods. 
Deterministic typically consider the microsample represents the true correlation structure among the attributes. These methodologies tries to expand microsamples by fitting them to a aggregated marginal distribution. Introduced by \cite{Deming1940OnKnown}, Iterative Proportional Fitting (IPF) is one of the important key techniques used for population synthesis that combines the microsample data with aggregated marginal. In the review from \cite{E.Ramadan2020AModels}, the authors show the extensive study on IPF methods and it has been continuously developing by adding various extensions to deal with emerging issues. Several studies like \cite{Beckman1996CreatingPopulations, Zhu2014SyntheticMicrosimulation, Rich2018Large-scaleDenmark} have used variation of IPF algorithm to generate synthetic population. Typically, for IPF methods, all the attributes in the microsample have to be discrete and with limited categories. Fitting for large number of individual attributes quickly becomes computationally and memory-wise expensive. Because of the dependency on the original microsample, the IPF methods cannot approximate high-dimensional data. A common issue that comes with high-dimensional data is the problem of zero-cells, which in addition to rendering of sparse sample may also lead to convergence and division by zero problems described by \cite{Choupani2016PopulationResearch}.

In recent years, probabilistic-based simulation methods have gained momentum, offering more robust solutions to the challenges faced by deterministic methods. Research by \cite{Borysov2019HowSynthesis} emphasizes that simulation-based methods provide a systematic way of interpolating data. Even if specific agents do not exist in the original data, it may still be possible to sample these specific agents by combining agents in the original data. These methods excel in addressing high-dimensional problems, offering better scaling properties and fulfilling the need for more detailed populations. One of the first model that uses a probabilistic-simulation framework for population synthesis was introduced by \cite{Farooq2013SimulationSynthesis} where they employed a Markov Chain Monte Carlo (MCMC) algorithm based on Gibbs sampling to draw from a partial joint distribution of data, simulating draws from the original distribution. Subsequently, \cite{Sun2015ASynthesis} utilized a Bayesian network to model the joint distribution function for multiple individual attributes. While these methods generally outperform conventional deterministic models, they may encounter zero-cell problems, especially when tested on larger datasets.

The emergence of Deep Generative Models (DGM) in the machine learning community has introduced new possibilities for population synthesis. \cite{Borysov2019HowSynthesis} introduced a Variational Autoencoder (VAE) model to synthesize a population based on Danish Trip Diary. They compared the model against conventional algorithms like IPF and other generative models like Gibbs sampling and Bayesian Network. In their experiments, they found that the VAE model was able to address the problem of sampling zeros by generating agents that are virtually different from those in the original data but have similar statistical properties.

\newpage
Notably, Generative Adversarial Networks (GANs), introduced by \cite{Goodfellow2014GenerativeNets}, have been applied to population synthesis. Initially designed for image data, GAN models demonstrate the potential to learn high-dimensional features using neural networks and produce high-quality synthetic data. This has led to the utilization of GAN models in synthesizing not only images but also music, text, and structured tabular data. In recent studies, WGANs based models have been used extensively for generating tabular data. \cite{Walia2020SynthesisingWCGAN-GP} present the WGAN model using gradient penalty to produce tabular data that is indistinguishable from real data. \cite{Baowaly2019SynthesizingNetworks.} proposed an two version of WGAN model  for generating binary synthetic long-form electronic health related data. The authors claim that the improved models can generate more realistic-looking synthetic data that can be used to train other machine learning models better than previous proposed models. \cite{Xu2019ModelingGAN} presents a new model called CTGAN which uses a conditional generator based in WGAN and present itself as the new state-of-art tabular generation model. 

In the domain of population synthesis, there are some research that have used WGAN. \cite{Garrido2020PredictionModelling} extended the research by \cite{Borysov2019HowSynthesis} and applied the WGAN model to Dutch trip diary data, comparing the models against the VAE as well as Bayesian models. In their study, the WGAN model outperformed the VAE model while producing a significantly lower number of structural zeros in the data. Later, \cite{Kim2023ASynthesis} introduced two new loss functions to the WGAN models with the aim of ensuring that the trained generator produces fewer structural zeros and more sampling zero data. They tested the models against the naïve WGAN model and VAE, showcasing that the new loss functions indeed help WGAN models produce significantly fewer structural zeros while maintaining a good level of sampling zero data. As highlighted in the previous section, none of the mentioned research is based on the microsamples that are incomplete, which is often the case in real world. Hence, there is a need for a population synthesis method that can impute the missing information in the microsamples.

There are many different methods to impute missing information in tabular data. In the research \cite{Emmanuel2021ALearning}, authors highlight many classical imputation methods that are based on machine learning techniques like KNNImpute, MICE, MissForest and SMOTE. With the popularity of generative models, many GAN based models have been proposed for data imputation. \cite{Yoon2018GAIN:Nets} proposed a model called Generative Adversarial Imputation Nets (GAIN) where a they used a hint vector to train a generator-discriminator model that can imputes the missing components conditioned on what is actually observed, and outputs a completed vector. \cite{Neves2022FromGeneration} improves upon the GAIN model by using WGAN as base and propose three different models called SGAIN, WSGAIN-CP, and WSGAIN-GP.

In this paper, we use the WGAN model from \cite{Kim2023ASynthesis} and include the ideas presented by \cite{Neves2022FromGeneration} to train the generator model with a mask vector that indicate the location of missing information in the training data. In this regard, the proposed method is different from \cite{Neves2022FromGeneration} as we do not aim to create a WGAN imputation model, rather just uses the techniques to handle training generator model with missing data. 

\section{Methodology}\label{Method}
In mathematical terms, we consider a population of agents $n=1,2, \dots, N$ with each agent defined by vectors of $K$ attributes i.e. variables with agent’s individual characteristics and household characteristics. This population of agents is a representation of the actual population of $N$ individuals. The population synthesis problem is concerned with estimation of joint distribution of synthetic population $\hat{P}(X)$ that approximate the true joint distributions of attributes across a real population $P(X)$.  The data for creating a model for $P(X)$ come in form of dis-aggregated data collected from survey data in form of microsample $X$, typically with a sample size of $M<N$. To represent the incomplete microsamples, $X_r$, we picked $q$ attributes and replaced $r\%$ of rows (selected at random), in $X$ with NaN values. For this study, the microsamples, $X$ and $X_r$ represents the training data for the propsoed WGAN model.

A deep generative model represents a category of machine learning models designed to generate novel data samples resembling a given dataset. The core objective is to comprehend and model the inherent patterns, structures, and statistical features embedded in the training data. The fact that neural networks can approximate any function makes them a natural choice for the population synthesis problem, that is, the approximation of the function $P(X)$. This paper primarily focuses on the WGAN with gradient penalty (WGAN-GP), a generative model based on the framework introduced by \cite{Goodfellow2014GenerativeNets} and subsequently enhanced by \cite{Gulrajani2017ImprovedGANs}. The WGAN-GP model generates a synthetic population by transforming a random generated numbers from $K$-dimensional standard normal latent variable, $Z$. The WGAN-GP aims to generate output such that $G(Z)\to \hat{P}(X)$ such that they are consistent with actual population, $P(X)$.

For training, the generator network, $G(Z)$, is initiated with a draw from a latent variable $Z$. The draw is then transformed in such a way that the output has the same dimensions and shape as the real data. The second network, the discriminator network $D(X)$, receives an observation. This can either be from real data or from the generator $G(Z)$. The objective of the discriminator is to tell whether the information it receives comes from the real data or not. The training process continues until the $D$ network is no longer able to distinguish between generated and synthetic data. The pseudo-code of training of the WGAN-GP is presented in \textbf{\textit{Algorithm}} \ref{alg:wgan-gp-miss}.

The learning process in the model is based on $G(Z)$ and $D(X)$ playing this adversarial game based on the loss function given by, 
\begin{align}
\label{eq:loss}
\mathscr{L} = \mathscr{L}_D + \mathscr{L}_G + \lambda_{gp}*\mathscr{L}_{GP} + \lambda_{bd}*R_{BD} + \lambda_{ad}*R_{AD}
\end{align} 

where $\mathscr{L}_D$ is the discriminator loss, $\mathscr{L}_G$ is generator loss and $\mathscr{L}_{GP}$ is the loss from regularization by gradient penalty on the discriminator. $R_{BD}$ and $R_{AD}$ are the regularization terms to control the generation out-of-training samples. The term $\lambda_{gp}$, $\lambda_{bd}$ and $\lambda_{ad}$ are the model hyper-parameter which are manually selected to control the effect gradient penalty and two regularization terms, respectively.

For training data of batch-size $M$, the loss functions can be defined as,  

\begin{align}
\label{eq:loss_d}
\mathscr{L}_D = \frac{1}{M}\sum_{i=1}^{M}-D(X_i)+D(G(Z_i))
\end{align} 
\begin{align}
\label{eq:loss_g}
\mathscr{L}_G = \frac{1}{M}\sum_{i=1}^{M}-D(G(Z_i))
\end{align} 
\begin{equation}
\label{eq:loss_gp}
\begin{split}
\mathscr{L}_{GP} = \frac{1}{M}\sum_{i=1}^{M} (\left\| \nabla_{\tilde{X_i}} D(\tilde{X_i})  \right\|_2-1)^2 \\\\
\tilde{X_i} = \alpha \hat{X_i} + (1-\alpha) \hat{X_i}
\end{split}
\end{equation}

where, $\left\|.\right\|_2$ is the euclidean norm, $X_i$ and $\hat{X_i}$ are the real and generated data, respectively. $\alpha$ is a random number from a uniform distribution of $\alpha \in U[0,1]$. 

The two regularization in the equation are boundary distance regularization $R_{BD}$ and average distance regularization $R_{AD}$ are introduced by \cite{Kim2023ASynthesis} in order to promote sampling zero samples and restrict structural zero generation samples, respectively. These functions are expressed as, 

\begin{align}
\label{eq:loss_rbd}
R_{BD} = \frac{1}{M}\sum_{i=1}^{m} \min_{j\in \left \{ 1:N \right \}, i\in \left \{ 1:M \right \}} (DIST(\widehat{X}_i, X_j))
\end{align} 
\begin{align}
\label{eq:loss_rad}
R_{AD} = -\frac{1}{NM}\sum_{i=1}^{M}\sum_{j=1}^{N}  (DIST(\widehat{X}_i, X_j))
\end{align} 
\begin{align}
\label{eq:dist}
DIST(X_i, X_j) = \sqrt{(X_i - X_j)^2}
\end{align} 

where, $\widehat{X}_i$ are the generated data of size $M$, $X_j$ is the entire training data of size $N$. The $R_{BD}$ calculates the nearest distance from each generated data in the batch to entire $N$ training data and average them for $M$ generated batch, where as $R_{AD}$ computes the average for average distance to the entire training sample distribution of $M$ generated data.

\begin{algorithm}[htbp]
\caption{WGAN-GP with missing data}
\label{alg:wgan-gp-miss}
\begin{algorithmic}[1]
\Require Generator ($G$), Discriminator ($D$), Latent Variable ($Z$), Training Data ($X$), Epochs ($E$), Batch Size ($M$), $D$ updates per epoch ($n_d$), Gradient Penalty ($\lambda_{gp}$), Boundary distance ($\lambda_{bd}$), Average distance ($\lambda_{ad}$), mask ($Y$).
\Ensure Trained Generator $G$, Trained Discriminator $D$
\State Initialize $G$ and $D$
\For{Epoch $e = 1$ to $E$}
    \For{Batch $M$ in Data Loader}
        \For{Update $d$ = 1 to $n_d$}
            \State $Z \gets \mathscr{N}(0,1)$ of size $M$
            \State $\acute{X} \gets G(Z)$ 
            \State \textbf{Multiply with mask, $\widehat{X} \gets \acute{X} * Y$} 
            \State $D_{real} \gets D(X_m)$
            \State $D_{fake} \gets D(\widehat{X})$
            \State Wasserstein loss for the $D$, $\mathscr{L}_D \gets - D_{real} + D_{fake} $       
            \State Gradient Penalty $\mathscr{L}_{GP} \gets \lambda_{gp} * \mathbb{E} [(\left\| \nabla_{\tilde{X_i}} D(\tilde{X_i})  \right\|_2-1^2)]$
            \State Update $D$ parameters using $\mathscr{L} \gets \mathscr{L}_D + \mathscr{L}_{GP}$
        \EndFor    
        \State Generator Loss $\mathscr{L}_G \gets -D_{fake}$
        \State Regularization $R_{BD} \gets \min (DIST(\widehat{X}, X^S))$
        \State Regularization $R_{AD} \gets DIST(\widehat{X}, X^S)$
        \State Update $G$ parameters using $\mathscr{L} \gets \mathscr{L}_G + \lambda_{bd}R_{BD} + \lambda_{ad}R_{AD}$
    \EndFor
\EndFor
\end{algorithmic}
\end{algorithm}

\begin{figure}[ht]
\centering
{{\includegraphics[scale=0.8]{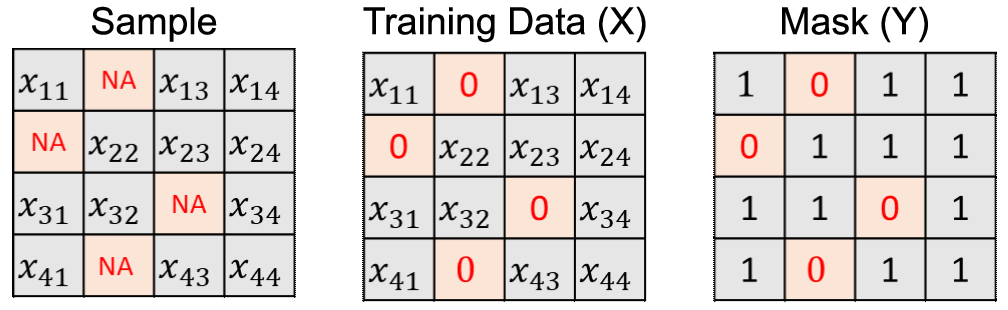}}}
\caption{Illustration showing an example of sample data with corresponding training data and mask. The missing values are represented as NA in the sample, which are replaced by 0 in training data and mask.}
\label{fig:missing_val}
\end{figure}

\newpage
In order to handle missing data, we used the masking approach presented by \cite{Neves2022FromGeneration}. During the training of the WGAN-GP, we introduce a new matrix called mask matrix as input. The mask, $Y$ as being the mask of training data, $X$, where a missing value in $X$ is represented by a zero and any non-missing value is represented by a one. Thus the mask is a binary matrix of same size as $X$. The biggest difference from the original WGAN training algorithm is in line 7 of the \textbf{\textit{Algorithm}} \ref{alg:wgan-gp-miss}, where the matrix generated by the generator, $G(Z)$ is multiplied by mask $Y$, before giving as input to discriminator $D$ to get score for fake samples. \textbf{\textit{Figure}} \ref{fig:missing_val} shows an example for the sample and its corresponding mask used during the training process. 

\newpage
\section{Case Study}\label{sec:CaseStudy}

\subsection{Travel Survey}\label{sec:Data}
The data for this study were obtained from the national travel behavior survey conducted by the Swedish Institute for Transport and Communications Analysis (SIKA), known as Riks-RVU 2005–2006 (\cite{RES0506}). This survey was carried out over a one-year period, from October 2005 to September 2006. It includes information on 41,000 individuals aged 6 to 84, selected through a stratified sampling process from the Swedish total population register. For the purposes of this analysis, we focus exclusively on the UPBD dataset within Riks-RVU, which contains detailed individual and household attributes. Each row in the survey represents a weighted sample that has been post-stratified based on strata defined by year, region, age, and sex. The regional classification is primarily at the county level, except for Stockholm County, where it is defined at the municipal level. Age groups are categorized as 6–14, 15–24, 25–44, 45–64, and 65–84 years.

The data were processed through the following steps to prepare them for model training: 1). All attributes were converted into categorical variables. 2). Missing values were imputed using domain knowledge to assign appropriate categories. 3). A full population dataset was generated by replicating each survey row according to its corresponding weight, thereby creating a synthetic population that represents the entire Swedish population in 2005. This replicated dataset comprises information on 8,227,341 individuals.

\begin{figure}[htbp]
\centering
{{\includegraphics[scale=0.5]{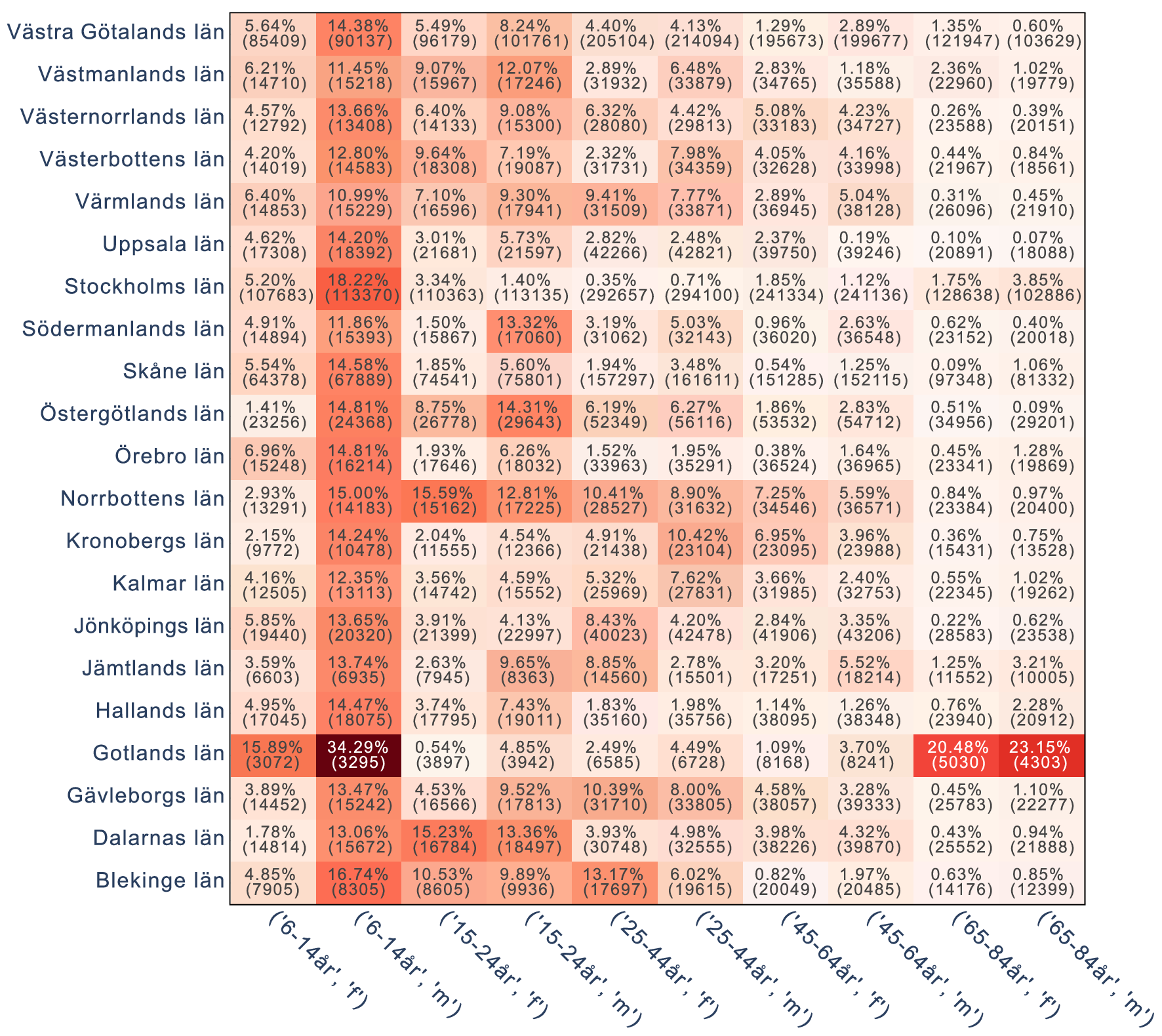}}}
\caption{Heatmap illustrating the disparity between the SCB and generated full population data for the marginal distribution of the population across all 21 counties in Sweden. Each cell displays the \% error and corresponding count from SCB 2006 (enclosed in brackets).}
\label{fig:pop_stats}
\end{figure}

To assess the accuracy of the generated full population in representing the actual Swedish population, we compared its marginal distributions with publicly available population statistics from Statistics Sweden (SCB) for the period January to December 2006. A heatmap visualization (see \textbf{\textit{Figure}} \ref{fig:pop_stats}) illustrates the count variations of females and males across different counties, comparing SCB data with the generated population. Visual inspection indicates that the synthetic population closely follows the trends observed in the SCB marginals for most regions and age groups. The most significant discrepancies appear in Götaland County and within the 6–14 age group. The elevated errors in Götaland County are likely due to its relatively smaller population size compared to the entire country. Differences in the 6–14 age group arise from divergent age group definitions: SCB reports for ages 5–14, whereas the generated population uses 6–14. Additional discrepancies may result from the slight misalignment in temporal coverage, as SCB data cover January to December 2006, while the Riks-RVU survey spans October 2005 to September 2006. Despite these minor inconsistencies, the generated full population dataset closely approximates the SCB marginals, supporting the conclusion that it provides a reliable representation of the actual population. Therefore, it can be confidently used as ground truth for subsequent analyses.

\subsubsection{Population Data}\label{sec:real_data}
In order to compare if the proposed methodology was able to successfully train WGAN models with missing information, we need a ground truth population data that does not contain any missing information. In that sense, we dropped  all rows containing any missing information from the generated full population dataset and retained only 17 distinct attributes pertaining to both individual and household characteristics. \textbf{\textit{Table}} \ref{tab:attributeTable} displays the list of attributes utilized in the project from the travel survey dataset, with 13 associated with individuals and 4 associated with households. This dataset is designated as \textit{h-population} serving as the reference dataset or "ground truth" for subsequent analysis. 

\begin{table}[htbp]
    \centering
    \caption{List of Attributes with Category proportions.}
    \label{tab:attributeTable}
    {{\includegraphics[scale=.8]{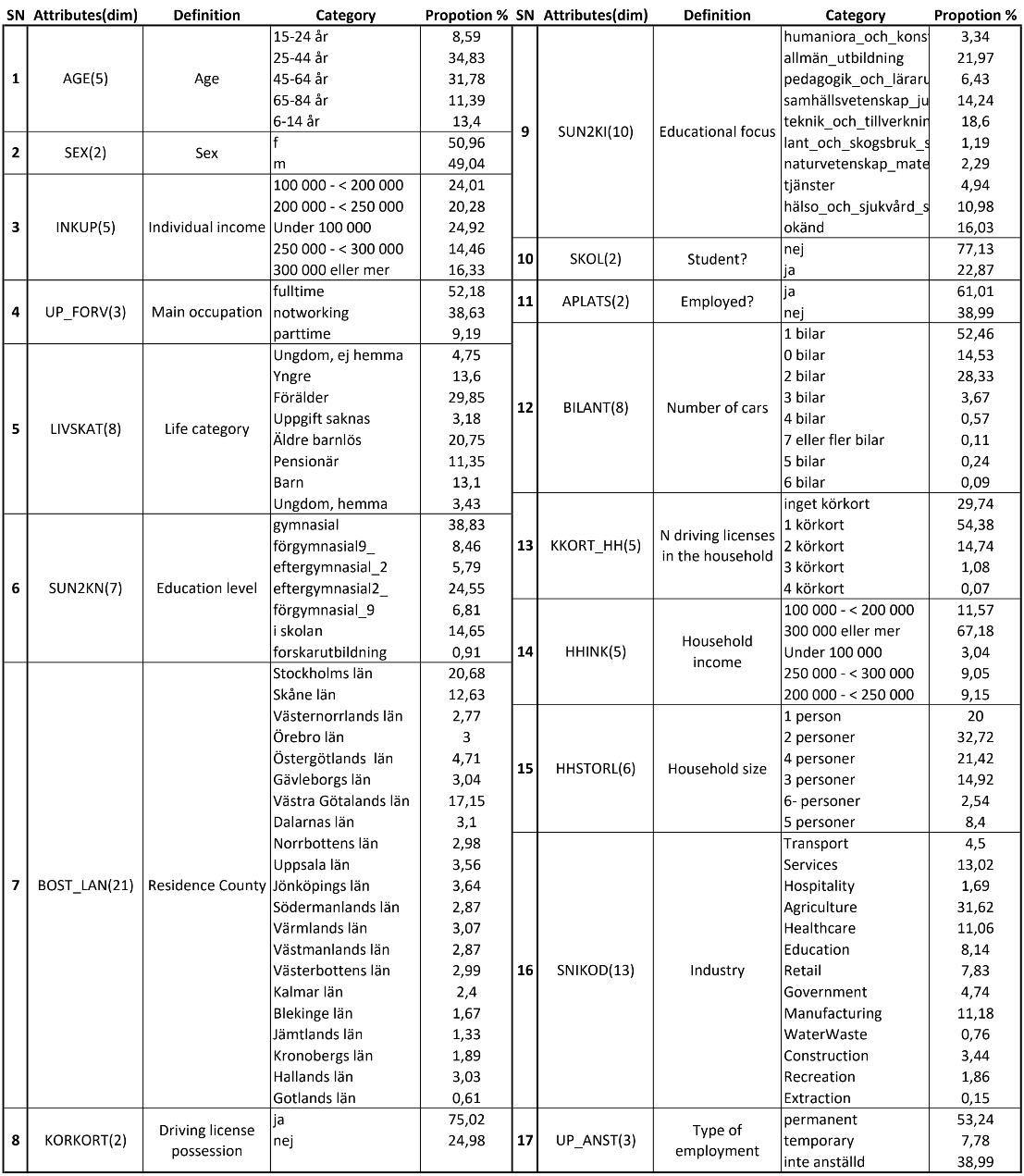}}}
\end{table}

\subsubsection{Training Data}\label{sec:training_data}
For training the proposed WGAN models, we create two different type of training dataset. First, we create a training set that is complete. We sample 10\% of the \textit{h-population} and deliberately remove certain unique category combinations from it. This refined dataset is labeled as \textit{h-nomis}. By eliminating certain combinations in \textit{h-nomis}, we address potential biases in structural zero and sampling zero rates arising from differences in the distributions of\textit{h-population} and \textit{h-nomis}.

Subsequently, we create multiple incomplete datasets. These datasets are derived from \textit{h-nomis} but include some attributes with missing information. These datasets simulates scenarios where information is either absent in microsamples or when multiple microsamples are merged, resulting in missing information on one or more attributes. The creation of incomplete datasets involves two steps: 1). randomly selecting a $q$ representing the number of attributes, and 2). introducing NaN values randomly to $r\%$ of each of the $q$ attributes in the \textit{h-nomis} dataset. These datasets are denoted as \textit{h-miss-q-r}, indicating the number of attributes and the proportion of rows with NaN values.

In the study, WGAN model trained with \textit{h-nomis} act as the "benchmark" model against which the other WGAN models trained on \textit{h-miss-q-r} will be analyzed and examined. In the end we generate synthetic poulation from each of the trained models and test them against \textit{h-population} for evaluating the performance of these models.  

\textbf{\textit{Table}} \ref{tab:data-stats} presents a comprehensive summary of the training data employed for the analysis, accompanied by their corresponding attributes.

\begin{table}[htbp]
\centering
\caption{Statistics on the datasets used for training and analysis.}
\label{tab:data-stats}
\resizebox{\textwidth}{!}{%
\begin{tabular}{clccccccc}
\textbf{SN}              & \multicolumn{1}{c}{\textbf{Dataset}} & \textbf{N Rows}              & \textbf{N Attributes}   & \textbf{N Categories}    & \textbf{\begin{tabular}[c]{@{}c@{}}N Unique \\ Combinations\end{tabular}} & \textbf{Missing Attributes}                                                                               & \textbf{\begin{tabular}[c]{@{}c@{}}Missing\\  Value (\%)\end{tabular}} & \textbf{\begin{tabular}[c]{@{}c@{}}N Missing Rows \\ (\% of Total)\end{tabular}} \\ \hline
\multicolumn{1}{|c|}{1}  & \multicolumn{1}{l|}{h-population}    & \multicolumn{1}{c|}{5156896} & \multicolumn{1}{c|}{17} & \multicolumn{1}{c|}{107} & \multicolumn{1}{c|}{14811}                                                & \multicolumn{1}{c|}{-}                                                                                    & \multicolumn{1}{c|}{-}                                                 & \multicolumn{1}{c|}{-}                                                           \\ \hline
\multicolumn{1}{|c|}{2}  & \multicolumn{1}{l|}{h-nomis}         & \multicolumn{1}{c|}{477489}  & \multicolumn{1}{c|}{17} & \multicolumn{1}{c|}{107} & \multicolumn{1}{c|}{12811}                                                & \multicolumn{1}{c|}{-}                                                                                    & \multicolumn{1}{c|}{-}                                                 & \multicolumn{1}{c|}{-}                                                           \\ \hline
\multicolumn{1}{|c|}{3}  & \multicolumn{1}{l|}{h-miss-2-10}     & \multicolumn{1}{c|}{477489}  & \multicolumn{1}{c|}{17} & \multicolumn{1}{c|}{107} & \multicolumn{1}{c|}{12811}                                                & \multicolumn{1}{c|}{UP\_FORV, SUN2KN}                                                                     & \multicolumn{1}{c|}{10}                                                & \multicolumn{1}{c|}{\begin{tabular}[c]{@{}c@{}}90576 \\ (18.97\%)\end{tabular}}  \\ \hline
\multicolumn{1}{|c|}{4}  & \multicolumn{1}{l|}{h-miss-2-20}     & \multicolumn{1}{c|}{477489}  & \multicolumn{1}{c|}{17} & \multicolumn{1}{c|}{107} & \multicolumn{1}{c|}{12808}                                                & \multicolumn{1}{c|}{UP\_FORV, SUN2KN}                                                                     & \multicolumn{1}{c|}{20}                                                & \multicolumn{1}{c|}{\begin{tabular}[c]{@{}c@{}}171707 \\ (35.96\%)\end{tabular}} \\ \hline
\multicolumn{1}{|c|}{5}  & \multicolumn{1}{l|}{h-miss-2-30}     & \multicolumn{1}{c|}{477489}  & \multicolumn{1}{c|}{17} & \multicolumn{1}{c|}{107} & \multicolumn{1}{c|}{12803}                                                & \multicolumn{1}{c|}{UP\_FORV, SUN2KN}                                                                     & \multicolumn{1}{c|}{30}                                                & \multicolumn{1}{c|}{\begin{tabular}[c]{@{}c@{}}243406 \\ (50.98\%)\end{tabular}} \\ \hline
\multicolumn{1}{|c|}{6}  & \multicolumn{1}{l|}{h-miss-2-40}     & \multicolumn{1}{c|}{477489}  & \multicolumn{1}{c|}{17} & \multicolumn{1}{c|}{107} & \multicolumn{1}{c|}{12779}                                                & \multicolumn{1}{c|}{UP\_FORV, SUN2KN}                                                                     & \multicolumn{1}{c|}{40}                                                & \multicolumn{1}{c|}{\begin{tabular}[c]{@{}c@{}}305761 \\ (64.04\%)\end{tabular}} \\ \hline
\multicolumn{1}{|c|}{7}  & \multicolumn{1}{l|}{h-miss-3-10}     & \multicolumn{1}{c|}{477489}  & \multicolumn{1}{c|}{17} & \multicolumn{1}{c|}{107} & \multicolumn{1}{c|}{12808}                                                & \multicolumn{1}{c|}{\begin{tabular}[c]{@{}c@{}}UP\_FORV, SUN2KN,\\ SUN2KI\end{tabular}}                   & \multicolumn{1}{c|}{10}                                                & \multicolumn{1}{c|}{\begin{tabular}[c]{@{}c@{}}129287 \\ (27.08\%)\end{tabular}} \\ \hline
\multicolumn{1}{|c|}{8}  & \multicolumn{1}{l|}{h-miss-3-40}     & \multicolumn{1}{c|}{477489}  & \multicolumn{1}{c|}{17} & \multicolumn{1}{c|}{107} & \multicolumn{1}{c|}{12613}                                                & \multicolumn{1}{c|}{\begin{tabular}[c]{@{}c@{}}UP\_FORV, SUN2KN,\\ SUN2KI\end{tabular}}                   & \multicolumn{1}{c|}{40}                                                & \multicolumn{1}{c|}{\begin{tabular}[c]{@{}c@{}}374565\\ (78.44\%)\end{tabular}}  \\ \hline
\multicolumn{1}{|c|}{9}  & \multicolumn{1}{l|}{h-miss-4-10}     & \multicolumn{1}{c|}{477489}  & \multicolumn{1}{c|}{17} & \multicolumn{1}{c|}{107} & \multicolumn{1}{c|}{12807}                                                & \multicolumn{1}{c|}{\begin{tabular}[c]{@{}c@{}}UP\_FORV, SUN2KN, \\ SUN2KI, HHSTORL\end{tabular}}         & \multicolumn{1}{c|}{10}                                                & \multicolumn{1}{c|}{\begin{tabular}[c]{@{}c@{}}164094 \\ (34.37\%)\end{tabular}} \\ \hline
\multicolumn{1}{|c|}{10} & \multicolumn{1}{l|}{h-miss-4-40}     & \multicolumn{1}{c|}{477489}  & \multicolumn{1}{c|}{17} & \multicolumn{1}{c|}{107} & \multicolumn{1}{c|}{12154}                                                & \multicolumn{1}{c|}{\begin{tabular}[c]{@{}c@{}}UP\_FORV, SUN2KN, \\ SUN2KI, HHSTORL\end{tabular}}         & \multicolumn{1}{c|}{40}                                                & \multicolumn{1}{c|}{\begin{tabular}[c]{@{}c@{}}415914 \\ (87.10\%)\end{tabular}} \\ \hline
\multicolumn{1}{|c|}{11} & \multicolumn{1}{l|}{h-miss-5-10}     & \multicolumn{1}{c|}{477489}  & \multicolumn{1}{c|}{17} & \multicolumn{1}{c|}{107} & \multicolumn{1}{c|}{12805}                                                & \multicolumn{1}{c|}{\begin{tabular}[c]{@{}c@{}}UP\_FORV, SUN2KN, \\ SUN2KI, HHSTORL, SNIKOD\end{tabular}} & \multicolumn{1}{c|}{10}                                                & \multicolumn{1}{c|}{\begin{tabular}[c]{@{}c@{}}195286 \\ (40.90\%)\end{tabular}} \\ \hline
\multicolumn{1}{|c|}{12} & \multicolumn{1}{l|}{h-miss-5-40}     & \multicolumn{1}{c|}{477489}  & \multicolumn{1}{c|}{17} & \multicolumn{1}{c|}{107} & \multicolumn{1}{c|}{11070}                                                & \multicolumn{1}{c|}{\begin{tabular}[c]{@{}c@{}}UP\_FORV, SUN2KN, \\ SUN2KI, HHSTORL, SNIKOD\end{tabular}} & \multicolumn{1}{c|}{40}                                                & \multicolumn{1}{c|}{\begin{tabular}[c]{@{}c@{}}440523 \\ (92.26\%)\end{tabular}} \\ \hline
\end{tabular}%
}
\end{table}

\newpage
\subsubsection{Sampling zero and Structural Zero}\label{sec:sampling_structural_zero}
\begin{figure}[htbp]
\centering
{{\includegraphics[scale=0.6]{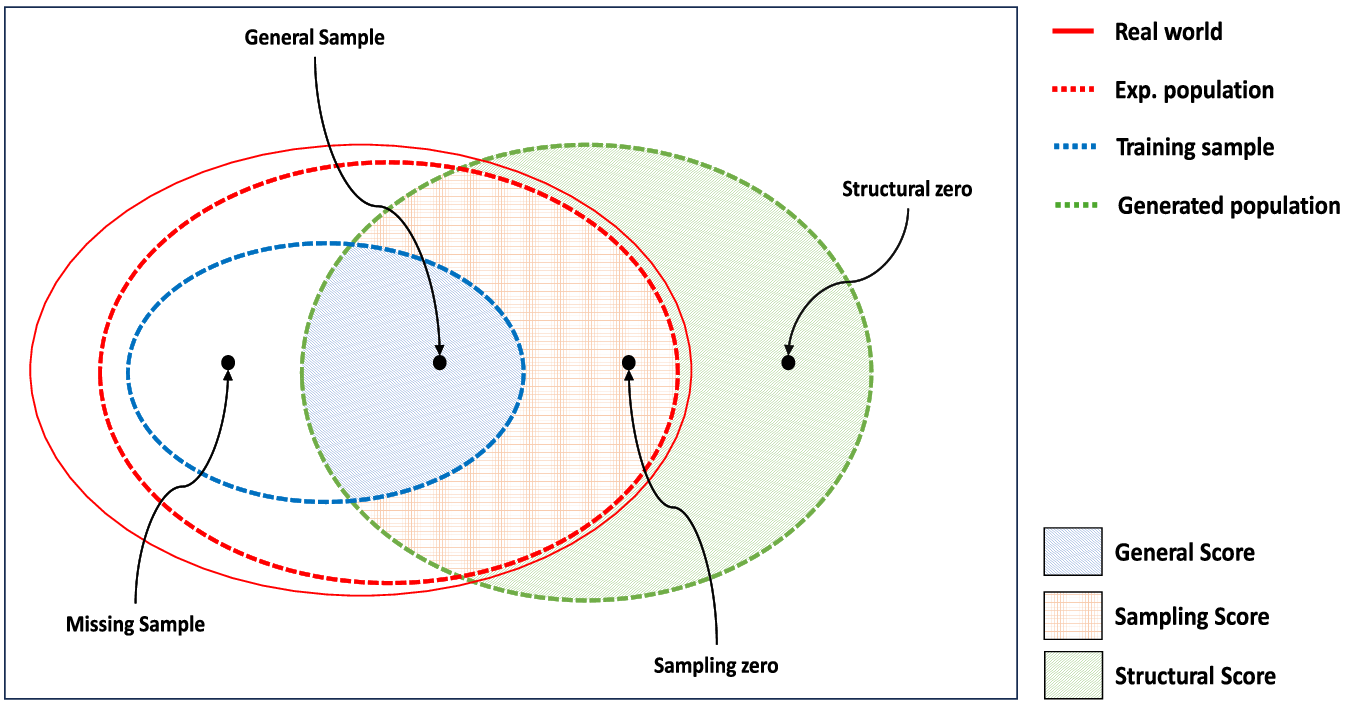}}}
\caption{Conceptual diagram showing the distribution of combination of categories and data types for our study.} \label{fig:typ_sample}
\end{figure}

As a WGAN based models can lean joint probability distributions of the attributes and hence can produce a variety of category combinations. In this section, we outline the different kinds of distribution of the category combination and data samples that can be produced from WGAN models. This concept is visually illustrated in the \textbf{\textit{Figure}} \ref{fig:typ_sample}. 

In the illustration, the black box represents the entirety of all possible combinations of the 107 categories in our study. Within this scope, the thin red circle outlines the distribution of  category combinations observed in the actual population, highlighted as "Real world" in the illustration. As we are using the population that is derived from a travel  survey, it may contain fewer category combinations when compared to actual population. The bold dashed RED circle illustrates the distribution of category combinations in the \textit{h-population} dataset, labeled as "Exp. population" in the illustration and lies within the Real world distribution. The training datasets, \textit{h-nomis} and \textit{h-miss-q-r}, represent a fraction of the \textit{h-population}, with certain combinations deliberately omitted from these samples. Consequently, this creates a distribution of category combinations smaller than the Exp. population, depicted as a bold dashed BLUE circle and referred to as the "Training Sample" in the illustration. Lastly, the distribution of category combinations from the synthetic population generated by trained WGAN models is depicted as a bold dashed GREEN circle, labeled as "Generated population" in the illustration. This distribution of the may intersect with the other distributions at various levels.

Following the classifications defined by \cite{Kim2023ASynthesis}, the synthetic population generated by WGAN models can be classified into four categories: general sample, sampling zero, structural zero, and missing sample. If a category combination is present in all distributions, it is termed a "general sample". The amount of general samples produced by the WGAN model is determined by the intersection of all distributions, depicted as the BLUE shaded region in the illustration. Categories that are absent in the Training Sample but feasible in the Exp. population and generated by the WGAN model are referred to as "sampling zero". The number of sampling zeros generated by the WGAN model is computed as the intersection of the Exp. population and Generated population, excluding the distribution of the Training Sample, and is illustrated as the RED shaded region. Next, categories generated by the WGAN model that are not part of either the Exp. population or Training Sample are categorized as "structural zero", shown as the GREEN shaded region. Lastly, samples present in the Training Sample but not generated by the WGAN model are termed "missing samples".

The objective of the population synthesis WGAN model is to achieve the highest level of intersection with both the Exp. population and Training Sample, thereby generating the maximum amount of general samples and sampling zeros while minimizing the number of structural zeros. As depicted in the illustration, it is desirable to have a large area of BLUE and RED shaded regions while reducing the area of the GREEN shaded region.

\newpage
\subsection{Model Evaluation and Discussion}\label{sec:Discussion}
To validate the effectiveness of our proposed approach for training WGAN with incomplete data, we initially utilized the \textit{h-nomis} dataset to optimize the hyperparameters of the WGAN model. Subsequently, we applied the same optimized parameters to train additional models using various \textit{h-miss-q-r} datasets. The hyperparameters specific to our scenario include number of layers and neurons for both the discriminator and generator, the dimension of the latent space vector, the learning rate, the regularization values for $\lambda_{bd}$,  $\lambda_{ad}$ and gradient penalty $\lambda_{gp}$. The models underwent training on a GPU cluster comprising four NVIDIA GeForce RTX 3080 units, each with 10GB of memory, for a total of 1000 epochs. The optimized model parameters are detailed in \textbf{\textit{Table}} \ref{tab:model_parms}.

Ultimately, the best-performing models were evaluated against the \textit{h-population} dataset, which serves as a ground truth for the real population. To assess the models, we generated synthetic populations named $G_{nomis}$ using the model trained on the \textit{h-nomis} dataset and use it as a benchmark to access other models. Another set of synthetic population, named $G_{miss-q-r}$, are generated using the model trained on the \textit{h-miss-q-r} dataset. All these synthetically generated populations consist of 5,156,896 data points.

\begin{table}[ht]
\centering
\caption{Parameters for the trained WGAN-GP models with regularization.}
\label{tab:model_parms}
\begin{tabular}{lc}
\multicolumn{1}{c}{\textbf{Parameter}}                   & \textbf{Value}                      \\ \hline
\multicolumn{1}{|l|}{Discriminator - N Layers}  & \multicolumn{1}{c|}{2}     \\ \hline
\multicolumn{1}{|l|}{Discriminator - N Neurons} & \multicolumn{1}{c|}{128}   \\ \hline
\multicolumn{1}{|l|}{Generator - N Layers}      & \multicolumn{1}{c|}{2}     \\ \hline
\multicolumn{1}{|l|}{Generator - N Neurons}     & \multicolumn{1}{c|}{128}   \\ \hline
\multicolumn{1}{|l|}{Latent Vector: N Neurons}  & \multicolumn{1}{c|}{128}   \\ \hline
\multicolumn{1}{|l|}{Learning Rate}             & \multicolumn{1}{c|}{0.01}  \\ \hline
\multicolumn{1}{|l|}{$\lambda_{gp}$}            & \multicolumn{1}{c|}{0.025} \\ \hline
\multicolumn{1}{|l|}{$\lambda_{bd}$}            & \multicolumn{1}{c|}{10}    \\ \hline
\multicolumn{1}{|l|}{$\lambda_{ad}$}            & \multicolumn{1}{c|}{1}     \\ \hline
\end{tabular}%
\end{table}

\subsubsection{Attribute-level Evaluation}
We first perform column level check on the WGAN generated synthetic populations to make sure that each attributes individually is able to follows statistical distribution of ground-truth data, \textit{h-population}. The attribute level check is done using three metrics - category coverage, TV complement and category adherence provided by \cite{Datacebo2024SDMetrics} library.

Category coverage  measures whether a attributes column in synthetic attribute covers all the possible categories that are present in ground-truth attribute. This metric first computes the number of unique categories, $C$, that are present in the ground-truth column $r$. Then it computes the number of those categories present in the synthetic attribute, $s$. It returns the proportion ground-truth categories that are in the synthetic data and is defined as, 
\begin{align}
\label{eq:cc_score}
score\_cc = \frac{C_s}{C_r}
\end{align} 

Total variation (TV) complement  metric computes the similarity of a ground-truth attribute vs. a synthetic attribute in terms of the column shapes i.e. the marginal distribution or 1D histogram of the column. This test computes the Total Variation Distance (TVD) between the ground-truth and synthetic attributes. To do this, it first computes the frequency of each category value and expresses it as a probability. The TVD statistic compares the differences in probabilities, as shown in \ref{eq:tvd}. 
\begin{align}
\label{eq:tvd}
\delta(R,S)=\frac{1}{2}\sum_{\omega\in\eta}^{}\left| R_\omega-S_\omega \right|
\end{align} 
Here, $\omega$ describes all the possible categories in a attribute, $\eta$. Meanwhile, R and S refer to the ground-truth and synthetic frequencies for those categories. The TV complement returns 1-TVD so that a higher score means higher quality and is give by, 
\begin{align}
\label{eq:tvs_score}
score\_tv = 1 - \delta(R,S)
\end{align} 

Category adherence metrics measures whether a synthetic attribute adheres to the same category values as the ground-truth data i.e. the synthetic population should not be inventing new category values that are not originally present in the ground-truth population. This metric extracts the set of unique categories, that are present in the ground-truth attribute, $Cr$. Then it finds the of data points of the synthetic data, $s$, that are found in the set $C$. The score is the proportion of these data points as compared to all the synthetic data points and is given by, 
\begin{align}
\label{eq:ca_score}
score\_ca = \frac{\left | s,s\in C_r \right |}{\left | s \right |}
\end{align}

\textbf{\textit{Table}} \ref{tab:single_eval} shows the metric scores for all attributes in the synthetic data generated by different WGAN models, tested against \textit{h-population} dataset. The metric score presented here demonstrate that the proposed WGAN training method successfully trains a model with incomplete data that closely approximates the performance of benchmark model. The visual inspection on the bar graphs, presented in \textbf{\textit{Appendix}} \ref{APX:bar_plots} for all 17 attributes for each of the dataset, further proves the analysis results. 

Upon deeper evaluation of \textbf{\textit{Table}} \ref{tab:single_eval}, it becomes apparent that certain $G_{miss-q-r}$ population exhibit slightly superior performance compared to benchmark $G_{nomis}$ population. Additionally, $G_{miss-q-r}$ population perform well for attributes with missing data - UP\_FORV, SUN2KN, SUN2KI, HHSTORL, SNIKOD and have similar metrics to $G_{nomis}$. A closer examination reveals that all populations (excluding $G_{miss-4-10}$ and$G_{miss-5-40}$) exhibit lower score\_cc for the "KKORT\_HH" attribute. This stems from the inability of these WGAN models to generate any sample with the "4-körkort" category. Similarly, the same trend is observed for the "BILANT" attribute, where WGAN models struggle to generate sample with the "6 billar" category. However, such performance levels are deemed acceptable for these trained WGAN models, considering the negligible share of "4-körkort" and "6 billar" categories in the actual population, as indicated in \textbf{\textit{Table}} \ref{tab:attributeTable}.
\begin{table}[btp]
    \centering
    \caption{Attribute level evaluation of all WGAN models against \textit{h-population}} \label{tab:single_eval}
    {\resizebox*{0.8\textwidth}{!}{\includegraphics{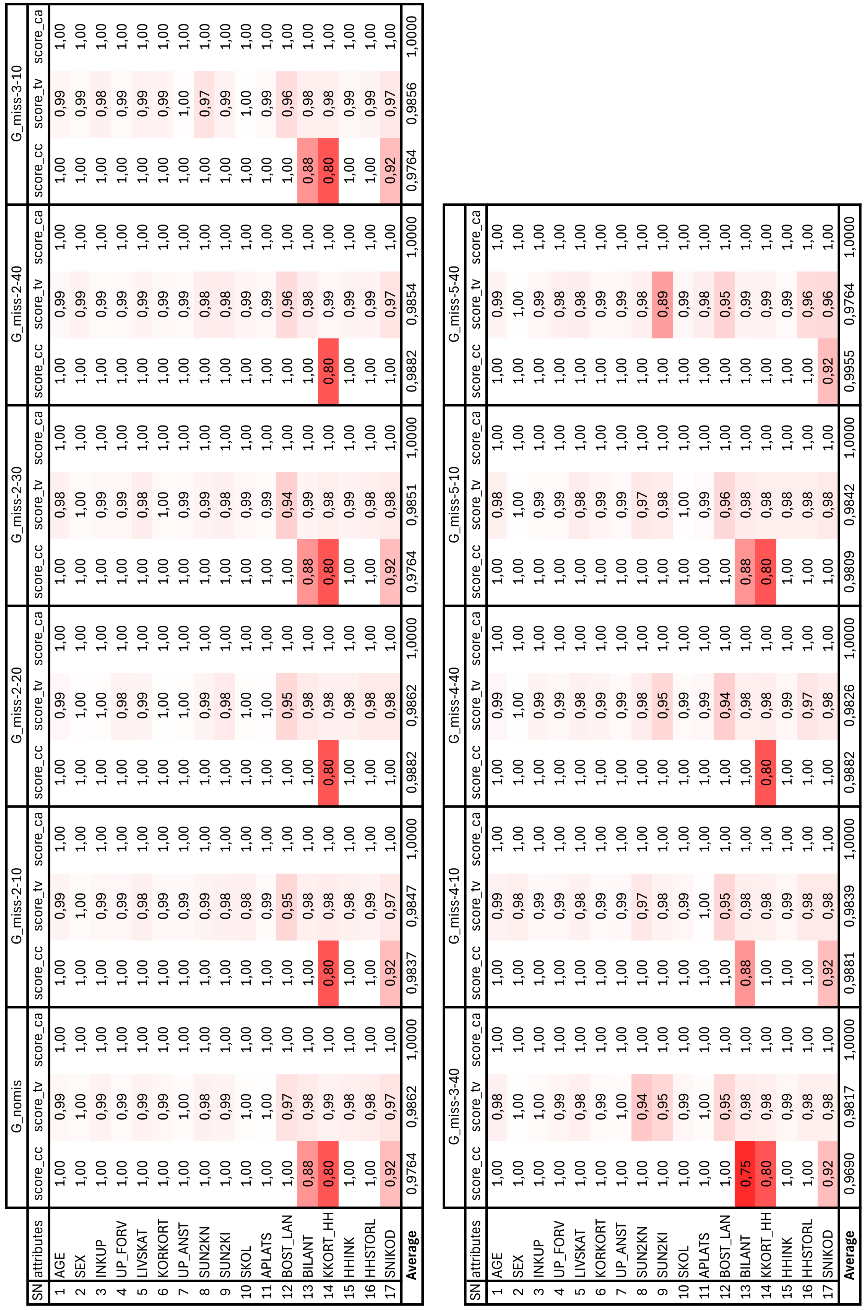}}}
\end{table}

\subsubsection{Higher Dimension Evaluation}\label{sec:high_srmse}
Following \cite{Kim2023ASynthesis, Garrido2020PredictionModelling, Borysov2019HowSynthesis}; we compare the categorical partial joint distribution of synthetic population generated by WGAN models. We used a standardized root mean square error (SRMSE) and coefficient of determination $R^2$ as metric for evaluation multi-dimensional distributions. The SRMSE is given by, 

\begin{align}
\label{eq:SRMSE}
SRMSE(\pi, \hat{\pi} | k)=\frac{RMSE}{\overline{\pi}}=\frac{\sqrt{\sum_{(i, j)}^{}(\pi_{(i, j)}-\hat{\pi}_{(i, j)})^2/N_{b}}}{\sum_{(i, j)}\pi_{(i, j)}/N_b} 
\end{align} 

where $\pi$ and $\hat{\pi}$ are k-joint categorical distribution of ground-truth and synthetic population, respectively. $N_b$ is the total number of possible category combinations and $k$ is the number of attributes in the joint. The $R^2$ score are computed using the categorical distribution for both the  ground-truth and synthetic population, and is given by, 
\begin{align}
\label{eq:R2}
R^2 = 1 - \frac{\sum_{(i, j)}^{}(\pi_{(i, j)}-\hat{\pi}_{(i, j)})^2}   {\sum_{(i, j)}^{}(\pi_{(i, j)}-\bar{\pi}_{(i, j)})^2} 
\end{align} 
where $\bar{\pi}$ is the mean of the k-joint categorical distribution of ground-truth population. 

We employ distinct subsets of attributes to encompass diverse features of the joint distribution. The evaluation is done for the following k-joint of attributes with total categories in brackets: 
\begin{itemize}
    \item 210-dimensional joint of AGE(5), SEX(2) and BOST\_LAN(21). 
    \item 16k-dimensional joint of UP\_FORV(3), SUN2KN(7), SUN2KI(10), HHSTORL(6) and SNIKOD(13).
    \item 7M-dimensional joint of AGE(5), SEX(2), BOST\_LAN(21), SUN2KI(10), SNIKOD(13), LIVSKAT(8), SUN2KN(7), and INKUP(5).
\end{itemize}

\textbf{\textit{Table}} \ref{tab:high-srmse} displays the outcomes derived from the SRMSE and $R^2$ metrics for all WGAN model across various k-joint levels.  We also present the counts for the count of unique category combinations that can be generated, are in the \textit{h-population} and are in each synthetic population, for all k-level joints. Comprehensive 45-degree charts for all models are provided in \textbf{\textit{Appendix}} \ref{APX:45deg_plots} for visual and qualitative evaluation.
 
The findings indicate that for all $G_{miss-q-r}$ population, their performance falls short of the benchmark $G_{nomis}$ population in terms of both SRMSE and $R^2$ values, across all k-dimensional joints. Consequently, as the amount of missing information in the training data increases, the SRMSE values (and corresponding $R^2$ values) tend to rise. This trend is particularly evident in the case of the 16k-dimensional joint, where attributes contain missing information in the dimensional joint. As the number of attributes with missing information grows, so does the SRMSE value, with $G_{miss-5-40}$ population exhibiting the highest error. In general, the $G_{miss-q-r}$ population only exhibits SRMSE and $R^2$ very close to the benchmark population $G_{nomis}$. Hence, it can be concluded that the proposed WGAN training method successfully trains a model with incomplete data. 

It is noteworthy that with an increase in the number of joint combinations, the SRMSE value rises (while $R^2$ decreases) across all models. This phenomenon occurs because the WGAN models generate a substantially larger number of category combinations compared to the ground-truth population. Some of these combinations are present within the ground-truth population, while others are not. We analyze this behavior by examining the count of sampling and structural zero data points produced by the models and evaluate the models' performance concerning the category combinations.

\begin{table}[hbtp]
    \centering
    \caption{Higher dimension evaluation for k-joint level distribution of attributes for all models.} \label{tab:high-srmse}
    {\resizebox*{\textwidth}{!}{\includegraphics{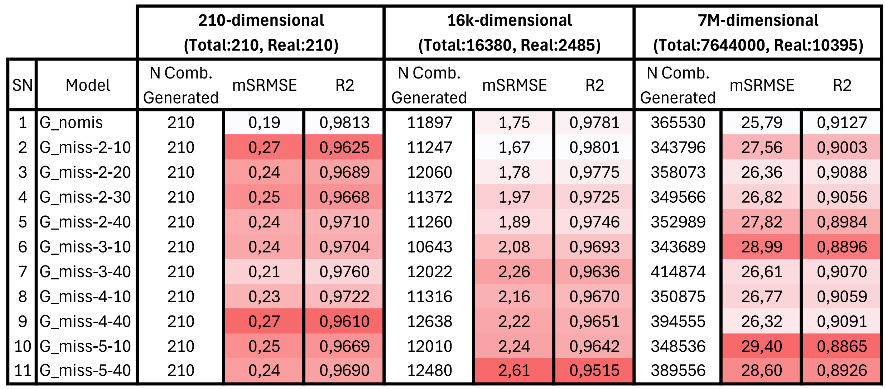}}}
\end{table}

\subsubsection{Sampling and Structural Zero Evaluation}
Based on the definitions provided in \textbf{\textit{Section}} \ref{sec:sampling_structural_zero}, we extracted general sample, sampling zero, and structural zero data for each generated population. This was done for k-joint attribute combination described in \textbf{\textit{Section}} \ref{sec:high_srmse} and at varied sampling levels. We computed the ratio of both general sample and sampling zero in the generated population in comparison to the combinations present in the ground-truth population (\textit{h-population}). We extracts the set of general samples ($GS$) or sampling zero ($SZ$)  that exists in \textit{h-population}. Then do the same for all WGAN generated population. The score is defined as, 

\begin{align}
\label{eq:score_gs}
score\_gs = \frac{GS_{generated}}{GS_{ground-truth}}
\end{align}

\begin{align}
\label{eq:score_ss}
score\_sz = \frac{SZ_{generated}}{SZ_{ground-truth}}
\end{align}

The ratio of structural zero is determined by calculating the total number of structural zero ($STZ$) instances generated by the WGAN models against all the unique combinations, $C$ produced by the WGAN model itself and is defined as, 

\begin{align}
\label{eq:score_stz}
score\_stz = \frac{STZ_{generated}}{C_{generated}}
\end{align}

In alignment with \cite{Kim2023ASynthesis}, we also implemented precision and recall to compare the models. Precision check weather the synthetic data generated new attributes combinations that still resembles the actual population. Recall measures the extent of over-fitting to the training sample. The value of precision and recall is given by,

\begin{align}
\label{eq:precision}
Precison = \frac{1}{C}\sum_{j=1}^{C}1_{ \hat{\pi}_j \in \pi}
\end{align} 

\begin{align}
\label{eq:recall}
Recall = \frac{1}{C}\sum_{j=1}^{C}1_{ \pi_j \in \hat{\pi}}
\end{align} 

\begin{figure}[hbtp]
   \centering
   \begin{subfigure}{\textwidth}
        \centering
        \includegraphics[scale=0.95]{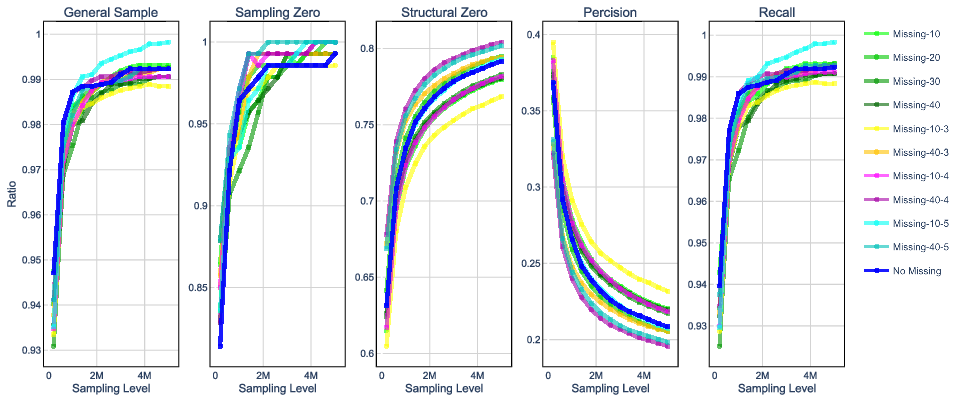}
        \caption{~16k-dim joint}
        \label{fig:syn_16k_sampling}
    \end{subfigure}

    \begin{subfigure}{\textwidth}
        \centering
        \includegraphics[scale=0.95]{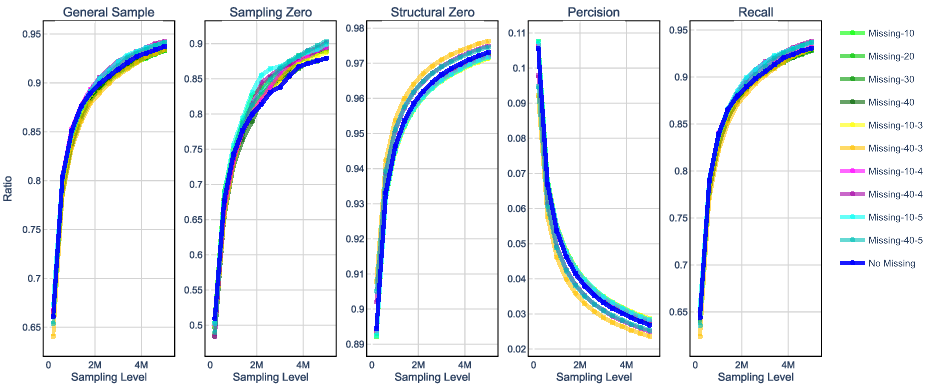}
        \caption{~7M-dim joint}
        \label{fig:syn_7M_sampling}
    \end{subfigure}

    \caption{Plots with the ratio of general sample, sampling zero, structural zero, precision and recall for 16k and 7M dimensional joint data at different sampling levels.}
    \label{fig:sampling_structural_zero}
\end{figure}

\textbf{\textit{Figure}} \ref{fig:sampling_structural_zero} displays the evaluation plots for all models, obtained from synthetic data generated from for 16k-dimensional and 7M-dimensional categorical joint, mentioned in \textbf{\textit{Section}} \ref{sec:high_srmse}.

In both presented plots, the metrics from all models exhibit similar or superior scores compared to the benchmark $G_{nomis}$ population. The metric score presented here demonstrate that the proposed WGAN training method successfully trains a model with incomplete data that closely approximates the performance of benchmark model.

With respect to the metric analysis, all models have successfully generated nearly all general samples and sampling zeros present in the ground-truth population. Generally, the number of sampling zeros and general samples generated increases as more data is sampled but decreases with the expansion of the k-dimensional space. For instance, when sampling from a 7M-dimensional distribution at a sampling rate of 5M data points, approximately $93.69\%$ \underline{+} $0.36\%$ of general samples and $89.42\%$ \underline{+} $0.8\%$ of sampling zeros are captured. At the same time, the number of structural zero also increases with the sampling rate as well as k-dimensional space, across all models. This is due to the fact, as the number of attribute dimensions is increasing, more and more combinations are possible within the data. The issue is that the total number of combinations that are present in the ground-truth population is very small compared to all possible unique combinations that are possible in an actual real world population. In this study, the \textit{h-population} only contains 10 395 unique combinations against the 7M possible combinations in the data which is only 0.13\% of all possible combinations. WGAN model struggles to restrict generation of structural zeros for such a small set of unique combinations in ground-truth data, especially for a high dimensional cases. This in fact is evident from the very low values for precision for all trained models, including the the base model $G_{nomis}$.  

In comparison to the results reported by \cite{Kim2023ASynthesis}, the precision values in our study are significantly lower. Possible explanation for this difference could be the unequal distribution of unique category combinations in the ground-truth data. Regarding \cite{Kim2023ASynthesis}, the dataset consists of 264,005 distinct combinations, while the \textit{h-population} used in the study only includes 14,811 distinct combinations. This dataset is 17 times smaller than the dataset mentioned in \cite{Kim2023ASynthesis}. Consequently, the precision and recall metrics rely on the quantity of distinct category combinations found in the actual data. Models trained on datasets with a greater number of category combinations will exhibit superior performance in terms of precision and recall metrics. This can be attributed to the capacity of WGAN models to generate a vast array of category combinations.

\newpage
\section{Conclusion}\label{sec:Conclusion}
The paper presents a novel method for population synthesis using WGAN models, which effectively train on incomplete data to ensure the synthetic population generated by the models is complete. This property of the model is especially useful when publicly accessible microsamples have missing information on one or more attributes; due to errors in data collection, privacy concerns resulting in data being withheld, or when information is missing during the merging of multiple microsamples. The proposed methodology utilizes a mask matrix to depict missing values in training data that allows the WGAN model to train on datasets that contain missing attributes.

The training method was validated using data from the Swedish national travel survey. We conducted a comparison between the benchmark model that was trained using complete data against models that were trained using data with different levels of missing information. The population generated from all the trained models was evaluated at the attribute-level and higher k-dimensional level to assess model's capability in generating sampling and structural zeros. For all the evaluation metrics, the results obtained from trained WGAN on incomplete data exhibit a high degree of similarity with the benchmark WGAN model trained on complete data. The validation results affirm the efficacy of the suggested training technique employing mask matrix in proficiently managing incomplete data, leading to synthetic populations that closely resemble the real population.

Upon closer examination of the evaluation results, it became evident that all trained WGAN models exhibited suboptimal performance on metrics relying on precision or SRMSE calculations, particularly in high-dimensional scenarios. This is attributed to the presence of numerous structural zeros generated by WGAN models, stemming from the limited number of unique category combinations in the travel survey data used for training in the study compared to the vast range of potential category combinations in real world. It is concluded that improved population data with a higher number of category combinations would enhance the metric results. Despite the significant number of challenges posed by structural zeros, all models excel in generating general samples and accurately sampling zeros, across various combinations of attributes and sampling levels.

The paper makes a substantial contribution to the field by providing a strong solution for population synthesis using incomplete data. This discovery presents new opportunities for future investigation, emphasizing the capacity of deep generative models to enhance the abilities of population synthesis, which is essential for agent-based models (ABMs) employed in transportation simulations and other fields.

\section{Future works}
Subsequently, the forthcoming course of action entails synthesizing the future population by employing the trained WGAN models. The conventional approach for this task involves employing Iterative Proportional Fitting (IPF) or Combinatorial Optimization (CO) techniques on a representative sample of the current year's population. This allows for the generation of a simulated population for a given future scenario that closely matches the desired distribution characteristics. An intriguing avenue for investigation would involve examining the feasibility of accomplishing this solely using an alternative deep learning model. Conditional Tabular GANs (CT-GAN) is a promising starting point for synthesizing data based on specified conditions. However, the current design of CT-GAN does not allow for conditioning on marginals. Further research is required to investigate the application of CT-GAN in generating synthetic populations for future scenarios, while considering marginal conditions.

\section{Acknowledgment}
The computations and data handling was enabled by the supercomputing resource Berzelius provided by National Supercomputer Centre at Linköping University and the Knut and Alice Wallenberg foundation. 

\bibliographystyle{apalike}
\bibliography{references}

\clearpage 
\newpage
\appendix

\newpage
\begin{landscape}
\section{Bar graphs for 17 attributes for different types of data}\label{APX:bar_plots}
\begin{table}[ht]
    \centering
    {{\includegraphics[scale=0.9]{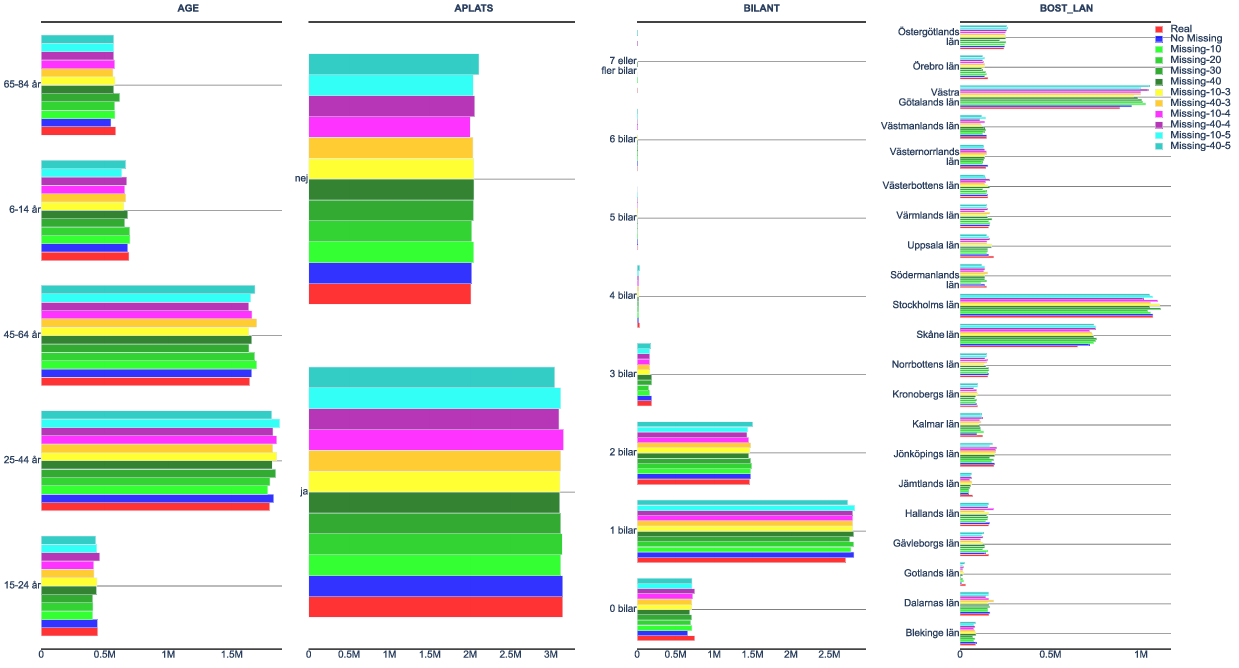}}}
\end{table}
\end{landscape}

\newpage
\begin{landscape}
\begin{table}[ht]
    \centering
    {{\includegraphics[scale=1]{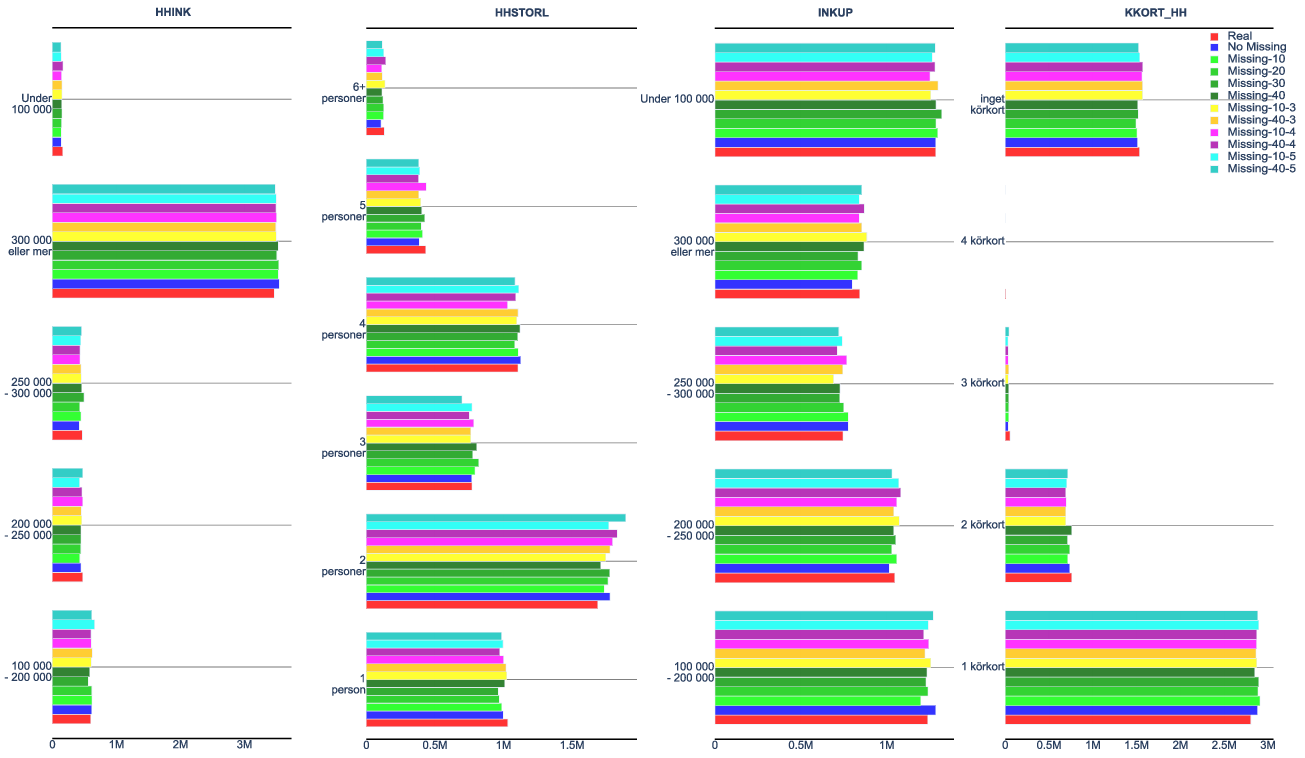}}}
\end{table}
\end{landscape}

\newpage
\begin{landscape}
\begin{table}[ht]
    \centering
    {{\includegraphics[scale=1]{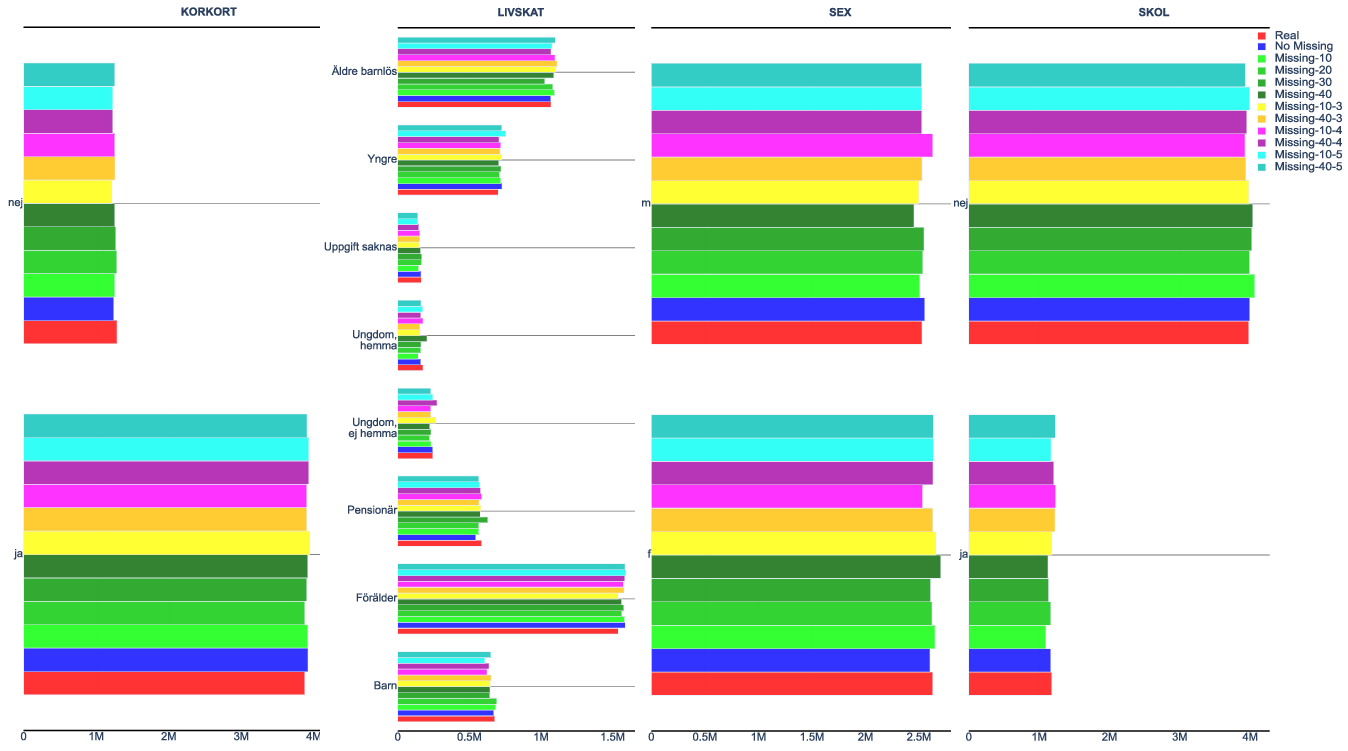}}}
\end{table}
\end{landscape}

\newpage
\begin{landscape}
\begin{table}[ht]
    \centering
    {{\includegraphics[scale=1]{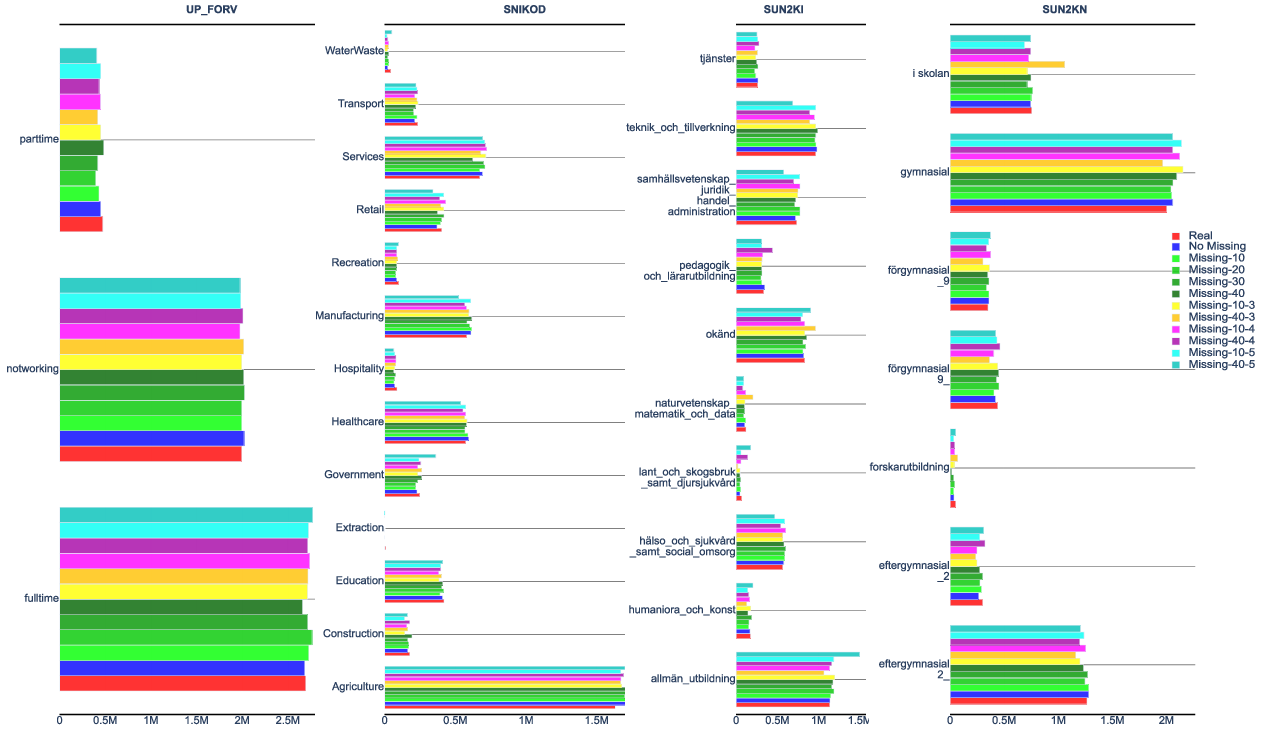}}}
\end{table}
\end{landscape}

\newpage
\begin{landscape}
\begin{table}[ht]
    \centering
    {{\includegraphics[scale=1]{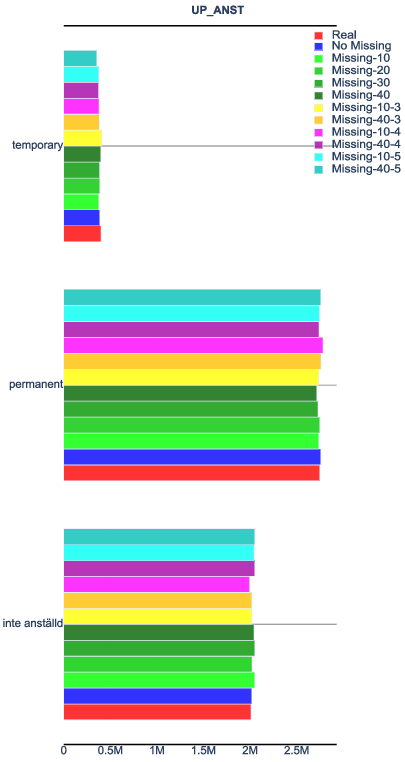}}}
\end{table}
\end{landscape}

\section{45-degree charts for higher dimensional qualitative assessment}\label{APX:45deg_plots}
\begin{figure}[ht]
\centering
{{\includegraphics[scale=0.65, angle=90]{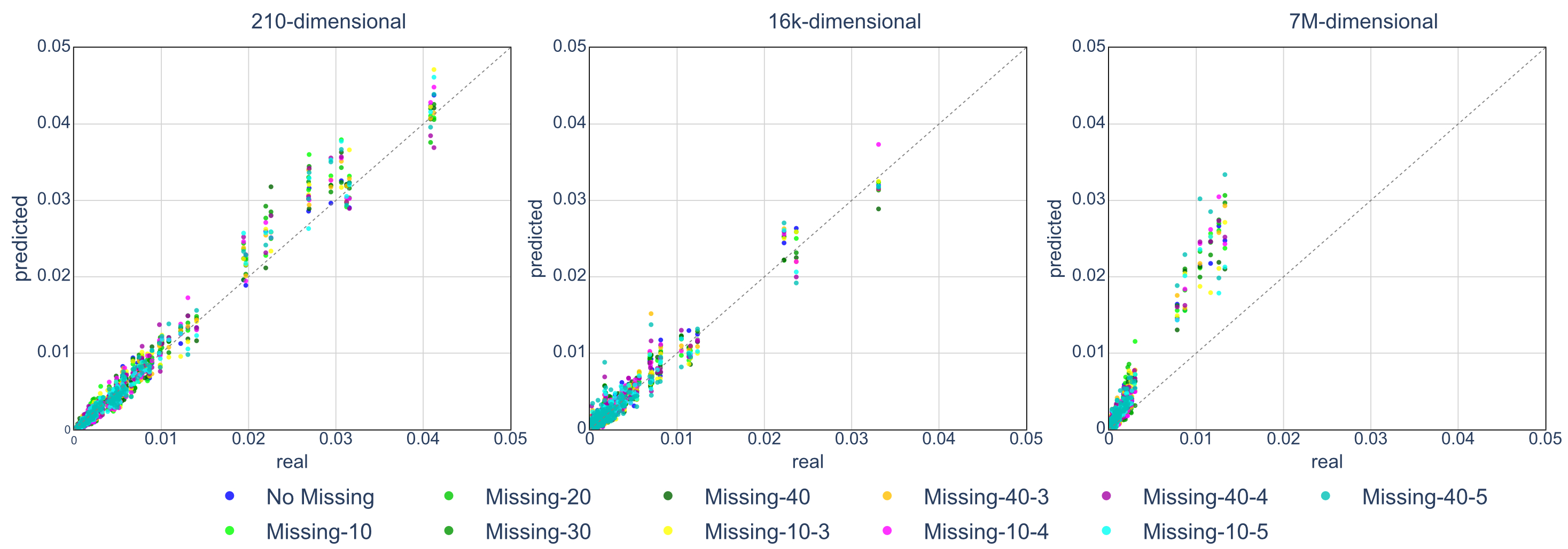}}}
\end{figure}

\end{document}

%% file: kth_thesis_style_papers.tex
\documentclass[12pt]{article}
\usepackage[T1]{fontenc} 
\fontfamily{bch}\selectfont
\usepackage[skip=8pt plus1pt, indent=0pt]{parskip}
\usepackage{titlesec}
\usepackage{fancyhdr}
\usepackage{setspace}
\usepackage{lmodern}
\usepackage{caption}
\usepackage{subcaption}
\usepackage[none]{hyphenat}

\newlength\titleindent
\newlength\sectionskip
\titleformat{\section}  
  {\vspace{\sectionskip}\fontsize{24}{24}\sffamily\raggedright} 
  {\llap{\parbox[b]{\titleindent}{\thesection\hfill}}} 
  {0em} 
  {} 
  [] 
\titlespacing*{\section}{0pt}{0pt}{24pt plus 1pt}

\titleformat{\subsection}  
  {\fontsize{15}{15}\sffamily\bfseries\raggedright} 
  {\thesubsection} 
  {0.6em} 
  {} 
  [] 
\titlespacing*{\subsection}{0pt}{18pt plus 3pt minus 1pt}{10pt plus 1pt}

\titleformat{\subsubsection}  
  {\fontsize{13}{13}\sffamily\bfseries\raggedright} 
  {\thesubsubsection} 
  {0.6em} 
  {} 
  [] 
\titlespacing*{\subsection}{0pt}{24pt plus 3pt minus 1pt}{12pt plus 1pt}